\documentclass[journal]{IEEEtran} 

\IEEEoverridecommandlockouts                              

\usepackage{graphicx}
\usepackage{epsfig} 
\usepackage{mathptmx}
\usepackage{amsmath} 

\usepackage{amssymb}  
\usepackage{amsthm}
\usepackage{wasysym}
\usepackage[mathcal]{euscript}
\usepackage{mathrsfs}
\usepackage{bm}
\usepackage{bbm}
\usepackage{color}
\usepackage{lipsum}
\usepackage[ruled,vlined,linesnumbered]{algorithm2e}
\usepackage[colorlinks=true,linkcolor=black,anchorcolor=black,citecolor=black,filecolor=black,menucolor=black,runcolor=black,urlcolor=black]{hyperref}
\usepackage{etoolbox}
\usepackage[table,xcdraw,dvipsnames]{xcolor}
\usepackage{multirow}
\usepackage{mathtools}
\usepackage{algpseudocode}
\usepackage{xspace}
\usepackage{afterpage}

\usepackage{outlines}

\usepackage{enumitem}
\setenumerate[1]{label=\Roman*.}
\setenumerate[2]{label=\Alph*.}
\setenumerate[3]{label=\arabic*.}
\setenumerate[4]{label=\roman*.}

\usepackage{pifont}

\usepackage{booktabs}
\usepackage{cite}

\makeatletter
\patchcmd{\@makecaption}
  {\scshape}
  {}
  {}
  {}
\makeatother

\title{\LARGE \bf
Certifiably Safe Manipulation of Deformable Linear Objects via Joint Shape and Tension Prediction
}
\author{Yiting Zhang$^{1}$ and Shichen Li$^{2}$
\thanks{$^{1}$Yiting Zhang is with the Department of Robotics, University of Michigan, Ann Arbor, USA. Email: \texttt{yitzhang@umich.edu}.}
\thanks{$^{2}$Shichen Li is with the Department of Mechanical Science and Engineering, University of Illinois at Urbana-Champaign, USA. Email: \texttt{shichen8@illinois.edu}.}
}

\begin{document}

\newif\ifcommentson
\commentsontrue
\newcounter{PatrickCount}
\addtocounter{PatrickCount}{1}
\newcommand{\pat}[1]{\textcolor{OliveGreen}{\ifcommentson\textbf{(\thePatrickCount) PH}: (#1)\fi}\addtocounter{PatrickCount}{1}}

\newcounter{JonCount}
\addtocounter{JonCount}{1}
\newcommand{\jon}[1]{\textcolor{Maroon}{\ifcommentson\textbf{(\theJonCount) JM}: (#1)\fi}\addtocounter{JonCount}{1}}

\newcounter{BohaoCount}
\addtocounter{BohaoCount}{1}
\newcommand{\bohao}[1]{\textcolor{Orange}{\ifcommentson\textbf{(\theBohaoCount) BZ}: (#1)\fi}\addtocounter{BohaoCount}{1}}

\newcounter{ThoughtCount}
\addtocounter{ThoughtCount}{1}
\newcommand{\thought}[1]{\textcolor{Blue}{\ifcommentson\textbf{(\theThoughtCount) Outline}: (#1)\fi}\addtocounter{ThoughtCount}{1}}

\newcounter{ShreyCount}
\addtocounter{ShreyCount}{1}
\newcommand{\shrey}[1]{{\textcolor{RoyalBlue}{\ifcommentson\textbf{(\theShreyCount) SK}: (#1)\fi}\addtocounter{ShreyCount}{1}}}

\newcounter{RamCount}
\addtocounter{RamCount}{1}
\newcommand{\Ram}[1]{\textcolor{WildStrawberry}{\ifcommentson\textbf{(\theRamCount) RV-M}: (#1)\fi}\addtocounter{RamCount}{1}}

\newcounter{FixCount}
\addtocounter{FixCount}{1}
\newcommand{\fix}[1]{\textcolor{Purple}{\ifcommentson\textbf{(\theFixCount) FIX}: (#1)\fi}\addtocounter{FixCount}{1}}

\newtheorem{defn}{Definition}
\newtheorem{rem}[defn]{Remark}
\newtheorem{lem}[defn]{Lemma}
\newtheorem{prop}[defn]{Proposition}
\newtheorem{assum}[defn]{Assumption}
\newtheorem{ex}[defn]{Example}
\newtheorem{thm}[defn]{Theorem}
\newtheorem{cor}[defn]{Corollary}
\newtheorem{con}[defn]{Conjecture}
\newtheorem{problem}[defn]{Problem}

\providecommand{\R}{\ensuremath \mathbb{R}}
\providecommand{\IR}{\ensuremath \mathbb{IR}}
\providecommand{\N}{\ensuremath \mathbb{N}}
\providecommand{\Q}{\ensuremath \mathbb{Q}}
\newcommand{\unitcircle}{\mathbb{S}^1}

\newcommand{\regtext}[1]{\mathrm{\textnormal{#1}}}
\newcommand{\ol}[1]{\overline{#1}}
\newcommand{\ul}[1]{\underline{#1}}
\newcommand{\defemph}[1]{\emph{#1}}
\newcommand{\ts}[1]{\textsuperscript{#1}}

\newcommand{\comp}{^{\regtext{C}}}
\newcommand{\card}[1]{\left\vert#1\right\vert}
\newcommand{\proj}{\regtext{proj}}
\newcommand{\norm}[1]{\left\Vert#1\right\Vert}
\newcommand{\abs}[1]{\left\vert#1\right\vert}
\newcommand{\pow}[1]{\regtext{pow}\!\left(#1\right)} 
\newcommand{\diag}[1]{\regtext{diag}\!\left(#1\right)}
\newcommand{\eig}[1]{\regtext{eig}\!\left(#1\right)}
\newcommand{\union}{\bigcup}
\newcommand{\intersection}{\bigcap}
\newcommand{\trans}{^\top}
\newcommand{\inv}{^{-1}}
\newcommand{\pinv}{^{\dagger}}
\newcommand{\sign}{\regtext{sign}}
\newcommand{\expm}{\regtext{exp}}
\newcommand{\logm}{\regtext{log}}
\newcommand{\skw}{_{\times}}
\newcommand{\bigO}{\mathcal{O}}
\newcommand{\bdry}[1]{\regtext{bd}\!\left(#1\right)}
\renewcommand{\ker}[1]{\regtext{ker}\!\left(#1\right)}
\newcommand{\convhull}[1]{\regtext{CH}\!\left(#1\right)}

\newcommand{\lbl}[1]{_{\regtext{#1}}}
\newcommand{\lo}{\lbl{lo}}
\newcommand{\hi}{\lbl{hi}}

\newcommand{\emptyarr}{[\ ]}
\newcommand{\zeros}{\textit{0}}
\newcommand{\ones}{\textit{1}}
\newcommand{\eye}{\regtext{\textit{I}}}


\newcommand{\interval}[1]{[ #1 ]}
\newcommand{\iv}[1]{[ #1 ]}
\newcommand{\nom}[1]{#1}
\newcommand{\pz}[1]{\mathbf{#1}}
\newcommand{\pzgreek}[1]{\bm{#1}}
\newcommand{\PZ}[1]{\mathcal{PZ}\left(#1\right)}
\newcommand{\pzk}[1]{\pz{ #1 ; k }}
\newcommand{\pzi}[1]{\pz{ #1 }(\pz{T_i};\pz{K})}
\newcommand{\pzki}[1]{\pz{ #1 }(\pz{T_i};k)}
\newcommand{\pzjki}[1]{\pz{ #1 }_j (\pz{T_i};k )}
\newcommand{\pzjKi}[1]{\pz{ #1 }_j (\pz{T_i};K )}
\newcommand{\setop}[1]{{\mathrm{\textnormal{\texttt{#1}}}}}
\newcommand{\numop}[1]{{\mathrm{\textnormal{\texttt{#1}}}}}
\newcommand{\lb}[1]{\underline{#1}}
\newcommand{\ub}[1]{\overline{#1}}


\newcommand{\ith}{$i$\ts{th}}
\newcommand{\jth}{$j$\ts{th}}

\newcommand{\pzqi}{\pzi{q}}
\newcommand{\pzqAi}{\pzi{\qA}}
\newcommand{\pzqdi}{\pzi{\dot{q}}}
\newcommand{\pzqdai}{\pzi{\dot{q}_{a}}}
\newcommand{\pzqddi}{\pzi{\ddot{q}}}
\newcommand{\pzqddai}{\pzi{\ddot{q}_{a}}}
\newcommand{\pzqdesi}{\pzi{q_{d}}}
\newcommand{\pzqddesi}{\pzi{\dot{q}_{d}}}
\newcommand{\pzqdddesi}{\pzi{\ddot{q}_{d}}}
\newcommand{\pzqdeski}{\pzki{q_{d}}}
\newcommand{\pzqddeski}{\pzki{\dot{q}_{d}}}
\newcommand{\pzqdddeski}{\pzki{\ddot{q}_{d}}}
\newcommand{\pzui}{\pzki{u}}
\newcommand{\pzqki}{\pzki{q}}
\newcommand{\pzqAki}{\pzki{\qA}}
\newcommand{\pzqdki}{\pzki{\dot{q}}}
\newcommand{\pzuki}{\pzki{u}}

\newcommand{\pzqji}{\pzi{q_j}}
\newcommand{\pzqAji}{\pzi{\qA_j}}
\newcommand{\pzqli}{\pzi{q_l}}
\newcommand{\pzqdji}{\pzi{\dot{q}_j}}
\newcommand{\pzqdaji}{\pzi{\dot{q}_{a,j}}}
\newcommand{\pzqddji}{\pzi{\ddot{q}_j}}
\newcommand{\pzqddaji}{\pzi{\ddot{q}_{a,j}}}
\newcommand{\pzqdesji}{\pzi{q_{d,j}}}
\newcommand{\pzqddesji}{\pzi{\dot{q}_{d,j}}}
\newcommand{\pzqdddesji}{\pzi{\ddot{q}_{d,j}}}
\newcommand{\pzfdes}{\pzki{q_{d,j}}}
\newcommand{\pzqddesjki}{\pzki{\dot{q}_{d,j}}}
\newcommand{\pzqdddesjki}{\pzki{\ddot{q}_{d,j}}}
\newcommand{\pzuji}{\pzki{u_j}}
\newcommand{\pzqjki}{\pzki{q_j}}
\newcommand{\pzqdjki}{\pzki{\dot{q}_j}}
\newcommand{\pzujKi}{\pz{u}(\pzqAi, \nomparams, \intparams, \pzfi)}
\newcommand{\pzujki}{\pz{u}(\pzqAki, \pzfi)}
\newcommand{\pzFKjki}{\pz{FK_j}(\pzqki)}
\newcommand{\pzFOjki}{\pz{FO_j}(\pzqki)}
\newcommand{\pzFKjKi}{\pz{FK_j}(\pzqi)}
\newcommand{\pzFOjKi}{\pz{FO_j}(\pzqi)}
\newcommand{\pzfi}{\mathbf{f}\left ( \mathbf{T}_{\mathbf{i}} ; k\right )}
\newcommand{\pzfdesi}{\pzi{f_{d}}}
\newcommand{\pzfki}{\pzki{f}}
\newcommand{\pzfdeski}{\pzki{f_{d}}}

\newcommand{\pzpboundj}{\bm{\epsilon}_{\mathbf{p, j}}}
\newcommand{\pzvboundj}{\bm{\epsilon}_{\mathbf{v, j}}}
\newcommand{\pzfbound}{\bm{\epsilon}_{\mathbf{f}}}

\newcommand{\pzg}{g}
\newcommand{\pzv}{x}
\newcommand{\pze}{\alpha}
\newcommand{\pzn}{{n_g}}
\newcommand{\pzgi}{g_i}
\newcommand{\pzei}{\alpha_i}

\newcommand{\q}{q(t)}
\newcommand{\qd}{\dot{q}(t)}
\newcommand{\qdd}{\ddot{q}(t)}
\newcommand{\qa}{q_a(t)}
\newcommand{\qadot}{\dot{q}_a(t)}
\newcommand{\qaddot}{\ddot{q}_a(t)}
\newcommand{\qak}{q_a(t; k)}
\newcommand{\qakdot}{\dot{q}_a(t; k)}
\newcommand{\qakddot}{\ddot{q}_a(t; k)}
\newcommand{\qdes}{q_d(t)}
\newcommand{\qdesdot}{\dot{q}_d(t)}
\newcommand{\qdesddot}{\ddot{q}_d(t)}
\newcommand{\qdesk}{q_d(t; k)}
\newcommand{\qdeskdot}{\dot{q}_d(t; k)}
\newcommand{\qdeskddot}{\ddot{q}_d(t; k)}


\newcommand{\qj}{q_j(t)}
\newcommand{\ql}{q_l(t)}
\newcommand{\qdj}{\dot{q}_{j}(t)}
\newcommand{\qddj}{\ddot{q}_{j}(t)}
\newcommand{\qaj}{q_{a, j}(t)}
\newcommand{\qajdot}{\dot{q}_{a, j}(t)}
\newcommand{\qajddot}{\ddot{q}_{a, j}(t)}
\newcommand{\qdesj}{q_{d, j}(t)}
\newcommand{\qdesjdot}{\dot{q}_{d, j}(t)}
\newcommand{\qdesjddot}{\ddot{q}_{d, j}(t)}
\newcommand{\qdeskj}{q_{d, j}(t; k)}
\newcommand{\qdeskjdot}{\dot{q}_{d, j}(t; k)}
\newcommand{\qdeskjddot}{\ddot{q}_{d, j}(t; k)}

\newcommand{\nq}{n_q}
\newcommand{\nt}{n_t}
\newcommand{\nf}{n_f}
\newcommand{\Nq}{ N_q }
\newcommand{\Nt}{ N_t }
\newcommand{\Ne}{ N_e }

\newcommand{\err}{e}
\newcommand{\errdot}{\dot{e}}
\newcommand{\errdotj}{\dot{e}_j}
\newcommand{\errj}{e_j}
\newcommand{\errjdot}{\dot{e}_j}
\newcommand{\errddot}{\ddot{e}}
\newcommand{\roblyap}{V(\qA(t),\Delta)}
\newcommand{\robh}{h(\qA(t),\Delta)}
\newcommand{\robv}{v}
\newcommand{\robr}{r}
\newcommand{\robrj}{r_j}
\newcommand{\robw}{w}
\newcommand{\robrdot}{\dot{r}}
\newcommand{\roblyapmax}{\overline{V}(q, r)}
\newcommand{\roblyapdot}{\dot{V}(\qA(t))}

\newcommand{\robH}{H}
\newcommand{\robhmin}{\underline{h}(\qA(t),[\Delta])}
\newcommand{\robhdot}{\dot{h}(\qA(t))}
\newcommand{\roblevel}{V_M}
\newcommand{\robcoeff}{\gamma}
\newcommand{\robKinf}{\alpha}
\newcommand{\robgain}{\alpha_c}
\newcommand{\ultbound}{\sqrt{\frac{2 \roblevel}{\sigma_m}}}
\newcommand{\pbound}{\epsilon_p}
\newcommand{\vbound}{\epsilon_v}
\newcommand{\pboundj}{\epsilon_{p, j}}
\newcommand{\vboundj}{\epsilon_{v, j}}
\newcommand{\pboundvec}{E_p}
\newcommand{\vboundvec}{E_v}
\newcommand{\epvar}{x_{e_p}}
\newcommand{\evvar}{x_{e_v}}
\newcommand{\epvarj}{x_{e_{p, j}}}
\newcommand{\evvarj}{x_{e_{v, j}}}
\newcommand{\qA}{q_A}
\newcommand{\fbound}{\epsilon_f}
\newcommand{\efvar}{x_{e_f}}

\providecommand{\R}{\ensuremath \mathbb{R}}
\newcommand{\plan}{\lbl{p}}
\newcommand{\elapsed}{\lbl{e}}
\newcommand{\prev}{\lbl{prev}}
\providecommand{\tplan}{t\lbl{p}}
\providecommand{\tfin}{t\lbl{f}}

\newcommand{\zi}{z_i}
\newcommand{\zj}{z_j}
\newcommand{\rbf}{\mathbf{r}(t)}

\newcommand{\bM}{M}
\newcommand{\Mq}{M(\q, \Delta)}
\newcommand{\Mqdot}{\dot{M}(q(t), \Delta)}
\newcommand{\bMt}{\Tilde{M}}

\newcommand{\bC}{C}
\newcommand{\Cq}{C(\q, \qd)}
\newcommand{\Cqd}{C(\q, \qd, \Delta)}
\newcommand{\bCt}{\Tilde{C}}

\newcommand{\bG}{G}
\newcommand{\Gq}{G(\q)}
\newcommand{\Gqd}{G(\q, \Delta)}
\newcommand{\bGt}{\Tilde{G}}

\newcommand{\bfc}{\mathbf{c}}
\newcommand{\bfI}{\mathbf{I}}

\newcommand{\intparams}{[\Delta]}
\newcommand{\intparamsdb}{[\Delta]\lbl{db}}
\newcommand{\nomparams}{\Delta_0}
\newcommand{\trueparams}{\Delta}
\newcommand{\pzparams}{\pzgreek{\Delta}}

\newcommand{\intforce}{[f](t)}
\newcommand{\nomforce}{f_{0}(t)}
\newcommand{\trueforce}{f(t)}

\makeatletter
\newcommand{\smalloplus}{\mathbin{\mathpalette\make@small\oplus}}
\newcommand{\smallotimes}{\mathbin{\mathpalette\make@small\otimes}}


\newcommand{\nominal}{\tau(t) = \bM(q(t), \nomparams) \ddot{q}_a(t) + \bC(q(t), \dot{q}(t), \nomparams) \dot{q}_a(t) + \bG(q(t), \nomparams)}

\newcommand{\robust}{ v = (\kappa(t) + \|\rho([\Phi]) \| + \varphi(t)) r}

\newcommand{\controller}{ u(t;k) = \tau(t;k) - \robv(t;k)}

\newcommand{\homtrans}{H}

\newcommand{\bPhi}{\Phi}

\newcommand{\wdistlong}{w(\Delta)}
\newcommand{\wdist}{w}
\newcommand{\wdistlongi}{w_i(\Delta)}

\newcommand{\wdistinterval}{w}


\newcommand{\closedloop}{\bH(\q,\Delta)\dot{r} + \bC(\q, \qd, \Delta)r  = -v + \wdist }

\newcommand{\lambdamin}{\lambda_m}
\newcommand{\lambdamax}{\lambda_M}
\newcommand{\sigm}{\sigma_{m}}
\newcommand{\sigM}{\sigma_{M}}

\newcommand{\Hquad}{\hspace{0.5em}}

\newcommand{\FK}{\regtext{\small{FK}}}
\newcommand{\IK}{\regtext{\small{IK}}}
\newcommand{\FO}{\regtext{\small{FO}}}
\newcommand{\IO}{\regtext{\small{IO}}}
\newcommand{\FS}{\regtext{\small{FS}}}
\newcommand{\IS}{\regtext{\small{IS}}}
\newcommand{\FC}{\regtext{\small{FC}}}
\newcommand{\IC}{\regtext{\small{IC}}}
\newcommand{\ID}{\regtext{\small{ID}}}

\newcommand{\exact}{^{\regtext{exact}}}
\newcommand{\slice}{\textnormal{\texttt{slice}}}
\newcommand{\eval}{\textnormal{\texttt{eval}}}
\newcommand{\stack}{\textnormal{\texttt{stack}}}
\newcommand{\reduce}{\textnormal{\texttt{reduce}}}
\newcommand{\getCoeffValue}{\texttt{getCoeffValue}}
\newcommand{\timeint}{([0, T])}

\newcommand{\Aobs}{A_O}
\newcommand{\bobs}{b_O}
\newcommand{\hobs}{h\lbl{obs}}
\newcommand{\nObs}{n_\mathscr{O}}
\newcommand{\obsset}{\mathscr{O}}

\newcommand{\SO}{\regtext{\small{SO}}}

\newcommand{\kj}{k_j}
\newcommand{\Kj}{K_j}

\newcommand{\kscale}{\eta_1}
\newcommand{\kjscale}{\eta_{j, 1}}
\newcommand{\koffset}{\eta_2}
\newcommand{\kjoffset}{\eta_{j, 2}}
\newcommand{\kvar}{x_k}
\newcommand{\kjvar}{x_{k_j}}

\newcommand{\tvar}{x_t}
\newcommand{\tvari}{x_{t_{i}}}

\newcommand{\F}{\mathcal{F}}

\newcommand{\eh}{\hat{e}}

\newcommand{\tsum}{{\textstyle\sum}}

\newcommand{\ujt}{u_j(t)}


\newcommand{\pjt}{p_j(t)}
\newcommand{\Rjt}{R_j(t)}










\newcommand{\unsafeobs}{\lbl{obs}}
\newcommand{\unsafeself}{\lbl{self}}
\newcommand{\unsafejoint}{\lbl{lim}}
\newcommand{\jlim}{\lbl{lim}}
\newcommand{\selfidx}{I\self}

\newcommand{\qlim}{q_{j,\regtext{lim}}}
\newcommand{\dqlim}{\dot{q}_{j,\regtext{lim}}}
\newcommand{\ddqlim}{\ddot{q}_{j,\regtext{lim}}}
\newcommand{\ulim}{u_{j,\regtext{lim}}}

\newcommand{\hitj}{h_i^{t, j}}
\newcommand{\buf}{\lbl{buf}}
\newcommand{\slc}{\lbl{slc}}
\newcommand{\Aitj}{A_i^{t, j}}
\newcommand{\bitj}{b_i^{t, j}}
\newcommand{\hitself}{h_{i_1, i_2}^{t}}
\newcommand{\Aitself}{A_{i_1, i_2}^{t}}
\newcommand{\bitself}{b_{i_1, i_2}^{t}}

\newcommand{\hqim}{h_{q_i^-}}
\newcommand{\hqip}{h_{q_i^+}}
\newcommand{\hdqim}{h_{\dot{q}_i^-}}
\newcommand{\hdqip}{h_{\dot{q}_i^+}}
\newcommand{\hijoint}{h_{i, \regtext{lim}}}











\newcommand{\initq}{q_{d_0}}
\newcommand{\initqj}{q_{d, j_{0}}}
\newcommand{\initdq}{\dot{q}_{d_0}}
\newcommand{\initdqj}{\dot{q}_{d, j_{0}}}
\newcommand{\initddq}{\ddot{q}_{d_0}}
\newcommand{\initddqj}{\ddot{q}_{d, j_{0}}}

\newcommand{\costfunc}{\phi}




\newcommand{\iss}{_{i}^{i}}
\newcommand{\issm}{_{i-1}^{i}}
\newcommand{\issmu}{_{i}^{i-1}}
\newcommand{\issmi}{_{i-1, i}^{i}}
\newcommand{\issp}{_{i+1}^{i}}
\newcommand{\issmm}{_{i-1}^{i-1}}
\newcommand{\isspp}{_{i+1}^{i+1}}
\newcommand{\issa}{_{a, i}^{i}}
\newcommand{\issc}{_{c, i}^{i}}
\newcommand{\issma}{_{a, i-1}^{i}}
\newcommand{\isspa}{_{a, i+1}^{i}}
\newcommand{\issmma}{_{a, i-1}^{i-1}}
\newcommand{\issppa}{_{a, i+1}^{i+1}}

\newcommand{\jss}{_{j}^{j}}
\newcommand{\jssm}{_{j-1}^{j}}
\newcommand{\jssmu}{_{j}^{j-1}}
\newcommand{\jssmj}{_{j-1, j}^{j}}
\newcommand{\jssp}{_{j+1}^{j}}
\newcommand{\jssmm}{_{j-1}^{j-1}}
\newcommand{\jsspp}{_{j+1}^{j+1}}
\newcommand{\jssa}{_{a, j}^{j}}
\newcommand{\jssc}{_{c, j}^{j}}
\newcommand{\jssma}{_{a, j-1}^{j}}
\newcommand{\jsspa}{_{a, j+1}^{j}}
\newcommand{\jssmma}{_{a, j-1}^{j-1}}
\newcommand{\jssppa}{_{a, j+1}^{j+1}}

\newcommand{\lssmu}{_{l}^{l-1}}

\newcommand{\qgoal}{q\lbl{goal}}
\newcommand{\qstart}{q\lbl{start}}
\newcommand{\methodname}{{ARMOUR}\xspace}

\newcommand{\normrho}{||\rho([\Phi])||}
\newcommand{\normwmax}{||w_M||}
\newcommand{\wmax}{w_M}
\newcommand{\wmaxj}{w_{M, j}}

\newcommand{\timestep}{\Delta t}

\newcommand{\xgoal}{\mathbf{x}\lbl{goal}}
\newcommand{\xstart}{\mathbf{x}\lbl{start}}

\maketitle
\thispagestyle{plain}
\pagestyle{plain} 

\begin{abstract}

Manipulating deformable linear objects (DLOs) is challenging due to their complex dynamics and the need for safe interaction in contact-rich environments. Most existing models focus on shape prediction alone and fail to account for contact and tension constraints, which can lead to damage to both the DLO and the robot. In this work, we propose a certifiably safe motion planning and control framework for DLO manipulation. At the core of our method is a predictive model that jointly estimates the DLO’s future shape and tension. These predictions are integrated into a real-time trajectory optimizer based on polynomial zonotopes, allowing us to enforce safety constraints throughout the execution. We evaluate our framework on a simulated wire harness assembly task using a 7-DOF robotic arm. Compared to state-of-the-art methods, our approach achieves a higher task success rate while avoiding all safety violations. The results demonstrate that our method enables robust and safe DLO manipulation in contact-rich environments.

\end{abstract}
\section{Introduction}

Effective manipulation of deformable linear objects (DLOs)—such as cables, ropes, and wires—is essential for a wide range of applications, including wire harness assembly and surgical suturing~\cite{lv2022dynamic,zhou2020practical,cao2019novel}. Yet, DLO manipulation remains a fundamental challenge in robotics due to their high-dimensional configuration spaces and nonlinear, time-varying dynamics~\cite{yin2021modeling}. Moreover, in real-world applications, contact between the DLO and its environment is often inevitable and sometimes even necessary~\cite{zhu2019robotic}. Such interactions introduce additional safety risks, including overextension and potential damage to both the DLO and the robot~\cite{sanchez2021four}. These challenges underscore the need for a manipulation framework that integrates high-fidelity modeling with planning and control strategies equipped with safety guarantees.

Existing modeling approaches for DLOs primarily focus on predicting shape deformations and can be broadly categorized into physics-based and learning-based methods. Physics-based models, such as mass-spring systems~\cite{lloyd2007identification}, position-based dynamics~\cite{liu2023robotic}, and the finite element method~\cite{sin2013vega}, offer physical interpretability but often suffer from a trade-off between high computational cost and limited accuracy in dynamic scenarios. Learning-based methods leverage deep neural networks to predict DLO shapes and sometimes incorporate physics priors to enhance consistency~\cite{yan2020self,yang2021learning,wang2022offline,chen2024differentiable}. While these models have achieved promising results in shape prediction, they face two key limitations: they generalize poorly to environments with frequent contact, and they cannot assess whether the DLO will be overextended—an essential factor in ensuring safe manipulation. Similar challenges have also been observed in structured manipulation settings involving cluttered scenes~\cite{zhang2025xpg}. Furthermore, shape prediction alone is insufficient; accurate tension estimation is critical for preventing failure and enabling reliable execution in safety-critical tasks. In real-world manipulation scenarios, however, tension information is rarely directly observable and is often difficult to model due to frictional contact, geometric constraints, and limited sensing~\cite{suberkrub2022feel}. These challenges make it difficult to analytically infer internal tension from geometric information alone. However, learning-based methods have been effectively applied in related contexts to approximate unobservable physical quantities~\cite{longhini2023edo,li2025multi,li2025multi2}, suggesting their potential to address this gap.


\begin{figure}[t]
    \centering
    \includegraphics[width=0.98\columnwidth]{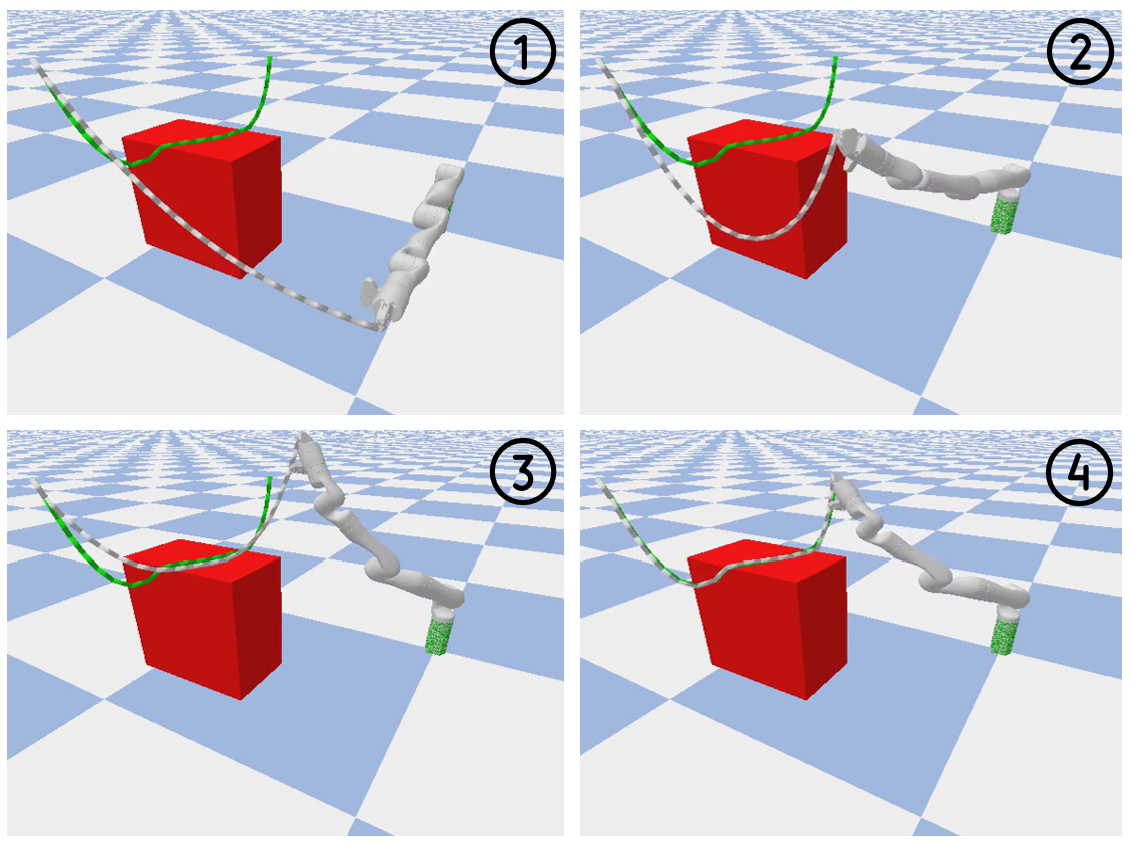}
    \caption{This paper introduces a certifiably safe framework for manipulating DLOs in contact-rich environments. At the core of our method is a predictive model that simultaneously estimates the future shape and tension of the DLO. The figure illustrates a robot arm manipulating a DLO (white) toward a goal configuration (green) in the presence of an obstacle (red). The manipulation is executed without causing collisions or overstretching, despite contact with the environment.}
    \label{fig:result_preview}
    \vspace*{-0.5cm}
\end{figure}

On the planning side, existing methods often rely on simplifying assumptions that compromise safety. Sampling-based approaches, such as probabilistic roadmaps (PRM)~\cite{saha2006motion} and rapidly-exploring random trees (RRT)~\cite{rodriguez2006obstacle, mcconachie2020manipulating}, as well as learning-based methods using supervised~\cite{mitrano2021learning} or reinforcement learning~\cite{lin2021softgym}, typically neglect contact-induced risks by assuming no overextension and no collisions between the robot and its environment. However, in real-world scenarios, a DLO can become trapped or overstretched, generating unexpected forces that alter the robot’s behavior and potentially lead to unanticipated collisions. These safety hazards underscore the need for a motion planning and control framework that not only accounts for DLO deformation but also certifiably enforces safety under physical constraints.

\begin{figure*}[t]
    \centering
    \includegraphics[width=2.0\columnwidth]{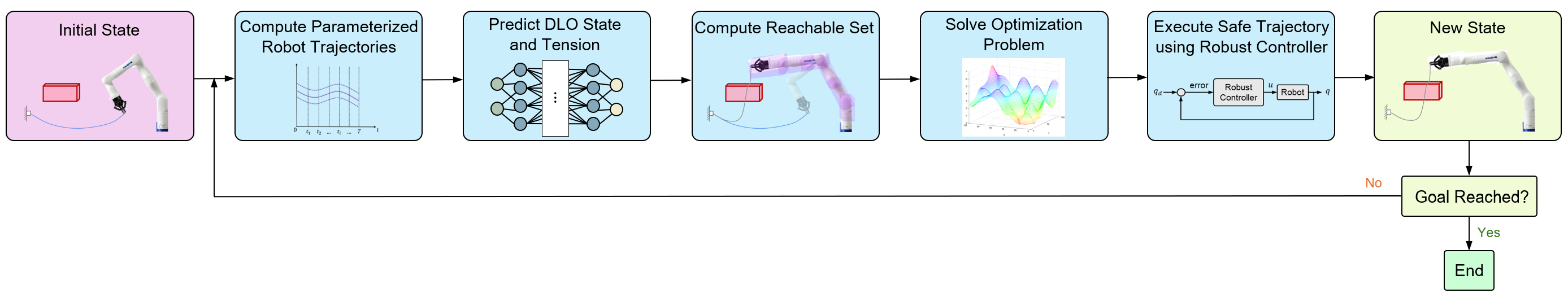}
    \caption{Overview of the proposed framework. Starting from an initial state, the robot generates a set of parameterized trajectories and uses a learned predictive model to estimate the resulting DLO shape and tension. These predictions are incorporated into a reachable set computation, enabling the identification of safe trajectories under contact-rich conditions. An optimization problem is then solved to select the best feasible trajectory, which is executed using a robust controller. The loop runs in a receding horizon fashion until the DLO reaches the desired goal configuration without collisions or overstretching.}
    \label{fig:method_overview}
    \vspace*{-0.5cm}
\end{figure*}

In this work, we propose a novel learning-based predictive model that jointly forecasts the future shape and tension of DLOs over extended time horizons. Building on this model, we develop a certifiably safe motion planning and control framework for DLO manipulation in complex, 3D contact-rich environments. The framework explicitly enforces safety constraints on DLO tension and robot motion through an online trajectory optimization process. This enables robust motion execution without inducing excessive tension or collisions. Experimental results in a simulated wire harness assembly task demonstrate substantial improvements in accuracy, safety, and robustness compared to existing baselines.

\section{Methods}
\label{sec:methods}

This paper addresses the problem of safe motion planning and control for DLO manipulation in contact-rich environments. We consider scenarios in which one end of the DLO is held by the robot’s end-effector, while the other end is either fixed to the environment or held stationary by another robot. The objective is to develop a framework that leverages predictions of DLO shape and tension to guide the robot in manipulating the DLO from an initial configuration to a desired one. The proposed approach aims to prevent both manipulator collisions and DLO overextension. An overview of the framework is shown in Fig.~\ref{fig:method_overview}.

\subsection{DLO Shape and Tension Prediction}
In this subsection, we present our approach to jointly predicting the shape and tension of a DLO during dynamic manipulation. We begin by introducing a discretized representation of the DLO. We then leverage the structure of a long short-term memory (LSTM) network to develop a model that predicts future DLO shape and tension. The model requires only the current DLO state and the planned future states of the manipulator as inputs.

\subsubsection{Discretized state representation}
\begin{figure}[t]
    \centering
    \includegraphics[width=0.98\columnwidth]{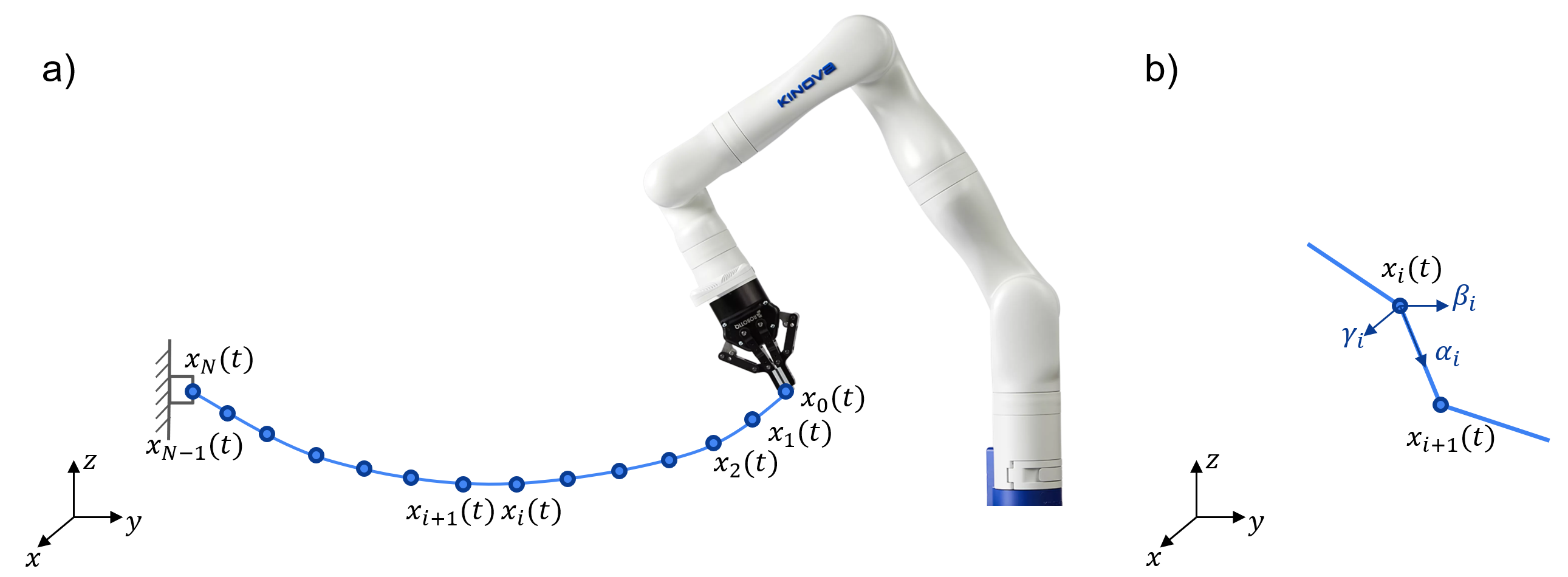}
    \caption{
    a) is an overview of the discretized configuration of DLO. Blue circles represent the discretized nodes of the DLO. b) shows the local coordinate of each node. For the $i^{\textrm{th}}$ node, its local coordinate is denoted by $(\alpha_{i}, \beta_{i}, \gamma_{i})$.}
    \label{fig:wire_disc}
    \vspace*{-0.2cm}
\end{figure}

Fig. \ref{fig:wire_disc} illustrates the discretized state representation for the DLO and shows the local coordinates for each discretized node. 
We define the state of the DLO at time $t$ as $\mathbf{x}(t)$. 
We discretize the DLO into $N$ small links with $N+1$ nodes so the state of entire DLO can be represented by the state of each node: $\textbf{x}(t) = [x_{0}(t), x_{1}(t), ..., x_{N}(t)]^{\top}$.
Each node's state can be explicitly denoted by its position in the world coordinates: $x_{i}(t)\in \mathbb{R}^{3}$ for $i = 0, 1, ..., N$.
Therefore, the DLO's state can be denoted by: $x\in \mathbb{R}^{N+1}$.
Neighbor DLO links are connected via two revolute joints ($\beta$ and $\gamma$ directions in local coordinate).

\subsubsection{Contact-aware modeling}
Given the state of the whole DLO at time $0$: $ \mathbf{x}_{0} =  \mathbf{x}(0)$, and the state of the end of DLO at time $t$: $x_{0}(t)$, we build a model to jointly predict the state of the whole DLO at time $t$:
\begin{equation}
    \textbf{x}(t) = f_{\lbl{NN, state}}(\mathbf{x}_{0}, x_{0}(t))
\end{equation}
and the tension at time $t = T$:
\begin{equation}
    f(t) = f_{\lbl{NN, tension}}(\mathbf{x}_{0}, x_{0}(t))
\end{equation}

\begin{figure}[t]
    \centering
    \includegraphics[width=0.98\columnwidth]{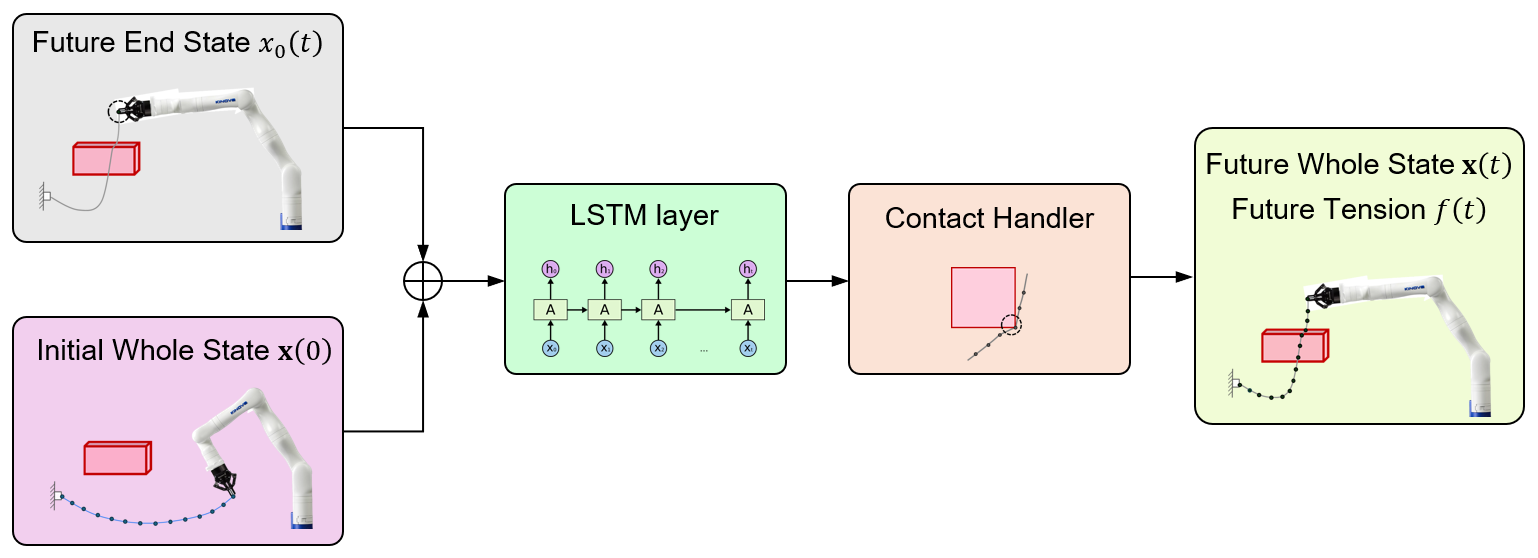}
    \caption{Architecture of the DLO shape and tension prediction model.
    }
    \label{fig:model_overview}
    \vspace*{-0.5cm}
\end{figure}

The model architecture is illustrated in Fig.~\ref{fig:model_overview}. It takes as input the initial full state of the system, $\mathbf{x}(0)$, and the future trajectory of the end-effector, $\mathbf{x}_0(t)$. An LSTM layer encodes temporal dynamics, while a contact handler explicitly accounts for interactions with obstacles. During contact, some nodes in the predicted DLO state may penetrate the obstacle model, which can degrade prediction accuracy. To address this, the contact handler solves a quadratic programming problem that projects the penetrated nodes onto the obstacle surface. This projection minimizes the displacement between each node’s original predicted position and its adjusted contact-consistent position. The model then outputs the predicted full state trajectory $\mathbf{x}(t)$ and the corresponding tension profile $f(t)$.

\subsection{Robot Model}
In the context of a serial robotic manipulator characterized by a configuration space $Q$ of dimension $\nq$ and a compact time interval $T \subset \R$, we define a trajectory for the configuration denoted as $q: T \to Q \subset \R^{\nq}$. 
The velocity associated with this configuration trajectory is represented by $\dot{q}: T \to \R^{\nq}$. 
Our assumption regarding the robot model is as follows:
\begin{assum}
The robot is composed of revolute joints, where the $j$\ts{th} joint actuates the robot's $j$\ts{th} link.
The robot has encoders that allow it to measure its joint positions and velocities.
The robot is fully actuated, where the robot's input $u: T \to \R^{\nq}$. 
\end{assum}

We further make the following assumption about the connection between the robot and the DLO:
\begin{assum}
The reaction force between the DLO and the robot is treated as an external force applied to the robot's end-effector. Its magnitude is equal to the tension in the DLO.
\end{assum}

The dynamics are represented by the standard manipulator equations \cite{spong2005textbook}:
\begin{equation}\label{eq:manipulator_equation}
\begin{aligned}
    M(q(t),\Delta)\ddot{q}(t) + C(q(t),\dot{q}(t),\Delta)\dot{q}(t) + G(q(t),\Delta) \\ = u(t)+J^{\top}(q(t))f(t)
\end{aligned}
\end{equation}
where $\Delta$ is the set of inertial parameters in the robot model, $\Mq \in \R^{\nq \times \nq}$ is the positive definite inertia matrix, $\Cqd \in \R^{\nq}$ is the Coriolis matrix, $\Gqd \in \R^{\nq}$ is the gravity vector, $u(t) \in \R^{\nq}$ is the input torque all at time $t$, $J(q(t)) \in\mathbb{R}^{3\times \nq}$ is the manipulator Jacobian, and $f(t)\in\mathbb{R}^{3}$ is the reaction force.

To reason about collision avoidance, we define the robot’s forward occupancy, which represents the space occupied by each link given the current joint configuration. Let $L_j \subset \mathbb{R}^3$ denote the volume of the $j$\ts{th} link in its local frame. Its forward occupancy at time $t$ is:
$\mathrm{FO}_j(q(t)) = p_j(q(t)) \oplus R_j(q(t)) L_j$, where $p_j(q(t))$ and $R_j(q(t))$ denote the position and orientation of joint $j$. The total arm occupancy is:
\begin{equation}
\mathrm{FO}(q(t)) = \bigcup_{j=1}^{\nq} \mathrm{FO}_j(q(t)).
\end{equation}
This is used to enforce collision-free motion in the planning framework.

\subsection{Online Trajectory Optimization}

Our online trajectory optimization builds upon the ARMOUR framework~\cite{michaux2023can}, which employs reachability analysis via polynomial zonotopes for safe motion planning under dynamic uncertainty. ARMOUR considers uncertainty in the robot’s inertial parameters and computes conservative reachable sets to ensure safety. We extend this approach by integrating predictions of DLO shape and tension into the optimization problem.

\subsubsection{Trajectory parameterization}
We assume without loss of generality that the control input and trajectory of a planning iteration begin at time $t = 0$ and end at a fixed time $\tfin$.
We denote the time that the planning and control framework identifies a new trajectory parameter as $\tplan$.
In each planning iteration, the framework chooses a desired trajectory to be followed by the arm. 
These trajectories are chosen from a continuum of trajectories, which each uniquely determinated by a \textit{trajectory parameter} $k \in K$ and are written as $q_{d}(t;k)$.
The set $K \subset \R^{n_k}$, $n_k \in \N$, is compact and represents a user-designed continuum of trajectories. 
In general, $K$ can be designed to include trajectories designed for a wide variety of tasks and robot morphologies \cite{kousik2019_quad_RTD,liu2024refine,michaux2023can}.
We assume that $\ddot{q}_d(\,\cdot\,; k)$ is a Lipschitz continuously differentiable function and $\dot{q}_d(\tfin; k) = \ddot{q}_d(\tfin; k) = 0$.
Finally, we define the desired configuration of the DLO in each planning iteration as $\mathbf{x}_{d} = (x_{0,d}, x_{1,d}, ..., x_{N,d})\trans$.

\subsubsection{Polynomial zonotope overapproximation}
The proposed framework leverages polynomial zonotopes to overapproximate parameterized trajectories and rigorously characterize the reachable sets of key variables during planning. Polynomial zonotopes offer a compact yet expressive representation of uncertainty over time and parameter space, making them well-suited for real-time robust trajectory optimization. This representation enables the propagation of uncertainty in both time $t$ and trajectory parameters $k$ through the system dynamics. Using the techniques from~\cite{michaux2023can}, we compute overapproximations of the robot’s joint state, velocity, control input, forward occupancy, and the DLO tension.

\subsubsection{Cost function}
We propose a novel cost function that incorporates the DLO’s configuration, in contrast to conventional approaches that rely solely on the robot’s configuration.
In each planning iteration, we choose a timestep $\Delta t$ and divide the compact time horizon $T$ into $n_{t} = \frac{T}{\Delta t}$ time subintervals.
Then we define $n_{t}+1$ time stamps: $t_{0}, t_{1}, ..., t_{n_{t}}$, where $t_{0} = 0$, $t_{n_{t}} = \tplan$, and $t_{i} - t_{i-1} = \Delta t$ holds true for every $i \in \left\{ 1, 2, ..., n_{t}\right\}$.
We define the accumulated distance between the wire-harness' state in the trajectory and the desired goal state as the cost function of the trajectory optimization problem:
\begin{equation}
\begin{aligned}
\setop{cost}(k) &= \sum_{i=1}^{n_{p}}\left\| \textbf{x}(t_{i}) - \textbf{x}_{d} \right\| 
\\ &= \sum_{i=1}^{n_{p}}\left\| f_{\lbl{NN, state}}(\mathbf{x}_{0}, x_{0}(t_{i})) - \textbf{x}_{d} \right\|
\\ &= \sum_{i=1}^{n_{p}}\left\| f_{\lbl{NN, state}}(\mathbf{x}_{0}, f\lbl{FK}(q(t_{i}; k))) - \textbf{x}_{d} \right\|
\end{aligned}
\end{equation}
where $n_{p}\in \left\{ 0, 1, 2,... ,n_{t}\right\}$, $t_{n_{t}} = T$, and $f\lbl{FK}(\cdot)$ is the function that calls the forward kinematics to compute the position of the robot's end-effector.

\subsubsection{Safety constraints}
We now describe the set of constraints that ensure the safety of any feasible trajectory, including a newly introduced constraint to prevent overstretching of the DLO. The trajectory must satisfy the robot's joint position, velocity, and actuation limits at all times. These constraints must hold for each joint throughout the entire planning horizon. Additionally, the robot must avoid collisions with any obstacles in the environment. To prevent the DLO from being overstretched, we define a maximum allowable tension threshold, denoted as $f\lbl{lim}$. 

\begin{assum}
Since the true DLO tension is predicted by a learned model, we account for potential prediction errors by introducing a certified error bound $\epsilon$, such that:
\[
\left\| f_{\text{true}}(t_{i}; k) - f(t_{i}; k) \right\| \leq \epsilon, \quad \forall t_{i}, k.
\]
Here, \( f_{\text{true}}(t_{i}; k) \) denotes the true DLO tension at time \( t_i \) under a trajectory parameterized by \( k \), and \( f(t_{i}; k) \) denotes the corresponding predicted tension from the learned model \( f_{\text{NN}, \text{tension}} \).
\end{assum}
Under this assumption, we impose the following certifiably safe constraint:
\begin{equation}
\left\| f(t_i; k) \right\| + \epsilon \leq f_{\text{lim}}, \quad \forall i \in N_t.
\label{eq:tension_safe}
\end{equation}
This ensures that even in the worst case, where the model underestimates the true tension by up to \( \epsilon \), the actual tension will not exceed the safety limit. In our implementation, \( \epsilon \) is estimated from the maximum residual observed on the test set.

\subsubsection{Online optimization problem}
Given the trajectory parameterization, polynomial zonotope overapproximation, cost function, and safety constraints defined above, the robot's trajectory for manipulating the DLO can be computed by solving the following optimization problem:
\begin{align}
    \label{eq:pz_optcost}
    &\underset{k \in K}{\min} &&\numop{cost}(k) \\
    \label{eq:pz_optpos}
    &&& \hspace*{-0.75cm} \pzqjki \subseteq [\qlim^-, \qlim^+]  &\forall i \in N_t, j \in N_q \\
    \label{eq:pz_optvel}
    &&& \hspace*{-0.75cm}\pzqdjki \subseteq [\dqlim^-, \dqlim^+]  &\forall i \in N_t, j \in N_q \\
    \label{eq:pz_opttorque}
    &&& \hspace*{-0.75cm}\pzujki \subseteq [\ulim^-, \ulim^+]  &\forall i \in N_t, j \in N_q \\
    \label{eq:pz_optposcon}
    &&& \hspace*{-0.75cm}\pzFOjki \bigcap O = \emptyset  &\forall i \in N_t, j \in N_q \\
    \label{eq:pz_optfor}
    &&& \hspace*{-0.75cm}\left\| \pzfi \right\| + \epsilon \leq f\lbl{lim} &\forall i \in N_t.
\end{align}

\subsection{Robust Controller}
The proposed framework adopts a robust, passivity-based controller to track the specified trajectories. This controller builds on a variant of the Recursive Newton-Euler Algorithm (RNEA) introduced in \cite{michaux2023can}. It conservatively accounts for uncertainties in the manipulator's dynamics due to uncertain inertial parameters and provides an upper bound on the worst-case tracking error. Notably, our controller also incorporates the external force exerted by the DLO, a novel consideration not addressed in prior work.

\section{Experiments}
\label{sec:experiments}
We evaluate the proposed method in a simulated warehouse environment. The experimental task involves using a Kinova Gen3 7-DOF robotic arm to assemble a wire harness onto an automotive engine. 

\subsection{Experiment Setup}
\subsubsection{Experimental environment}
We construct the simulation environment using PyBullet. The engine is modeled as a static cuboid obstacle placed on a flat surface. Its pose is defined by \((x_{Obs}, y_{Obs}, \theta_{Obs})\), where \(x_{Obs}\) and \(y_{Obs}\) specify the engine’s center coordinates in the global frame, and \(\theta_{Obs}\) denotes its orientation around the global \(Z\)-axis. The wire harness measures 1.20~m in length and weighs 120~g. The engine dimensions are 0.30~m (length), 0.50~m (width), and 0.50~m (height).

\subsubsection{Task description}
The task requires the robot to manipulate the wire harness safely along a planned trajectory from an initial state to a goal state. Initially, the wire harness is held in free space without contacting the engine. In the goal state, the middle section of the wire harness must be placed securely onto the engine. The simulation experiment consists of 100 randomized trials. In each trial, the engine is placed at a random location and orientation within the environment to evaluate the robustness of the proposed method under varying spatial configurations.

\subsubsection{Prediction model training process}
To train the prediction model, we generate 2,000 simulated trajectories. In each, the robot holds one end of the wire harness and executes a random action while the other end remains fixed. The robot and wire harness begin from randomized configurations with zero initial velocity and acceleration. Each trajectory spans 2 seconds, producing 4,000 seconds (approximately 66.7 minutes) of simulation data. We record the wire harness’s shape and tension throughout each trajectory.

The dataset is split into 1,500 trajectories for training, 300 for validation, and 200 for testing. The prediction model is trained using an L1 loss function and optimized with Adam at a learning rate of \(1\times10^{-4}\). Early stopping is applied based on validation loss to avoid overfitting.

\subsection{Baselines}
We compare our method against two baselines. The first is \textbf{Learning Where to Trust}~\cite{mitrano2021learning}, which uses a classifier to identify robot motions that might lead to trapping. However, it lacks certifiable safety guarantees for both the robot and the DLO. The second is \textbf{ARMOUR}~\cite{michaux2023can}, a certifiably safe planner that ensures robot safety but does not account for DLO-specific risks such as overstretching.

\subsection{Results}
The outcomes of 100 randomized trials are summarized in Tab.~\ref{tab:random_obstacles_results}. Each trial is categorized into one of four outcomes:
\textbf{Goal Reached}—the task is successfully completed within 20 motion steps;
\textbf{Failed without Violation}—the task is not completed within the limit, but no safety constraint is violated;
\textbf{Robot Collision}—the robot collides with the engine;
\textbf{DLO Overextension}—the wire harness exceeds its allowable tension threshold.

\begin{table}[t]
\caption{Results from 100 simulation trials.}
\centering
\resizebox{\columnwidth}{!}{%
\begin{tabular}{c|c c c c}
\multicolumn{1}{c|}{Method} & \shortstack{Goal \\ Reached} & \shortstack{Failed \\ w/o Violation} & \shortstack{Robot \\ Collision} & \shortstack{DLO \\ Overextension} \\ \hline
Learning Where to Trust & 38 & 30 & 29 & 3 \\ \hline
ARMOUR & 54 & 20 & 0 & 26 \\ \hline
Ours & 76 & 24 & 0 & 0 \\ \hline
\end{tabular}
}
\label{tab:random_obstacles_results}
\end{table}

Comparing ARMOUR with our proposed method highlights the importance of explicitly predicting DLO tension and incorporating it into the safety constraints during motion planning. While ARMOUR ensures robot safety, it lacks protection for the DLO, leading to 26 cases of overextension. In contrast, our method achieves both robot and DLO safety, with zero constraint violations. Although Learning Where to Trust shows fewer DLO overextensions than ARMOUR, it is overly conservative, successfully completing the task in only 38 out of 100 trials. This is largely due to its reliance on a binary classification model that labels motions involving DLO-environment contact as ``risky." However, such contact is often necessary to complete the task. The model’s inability to distinguish between harmful and acceptable contact leads to frequent conservative stops and task failures. Our method, by predicting the tension of the DLO in contact-rich environments, enables safer and more effective manipulation. It allows necessary interactions with the environment while preserving safety for both the robot and the DLO.

\section{Conclusion}
\label{sec:conclusion}

We presented a certifiably safe framework for manipulating DLOs in contact-rich environments. At its core is a predictive model that jointly estimates the shape and tension of the DLO, which is integrated into a trajectory optimization-based planner to enforce safety constraints on both the robot and the DLO during execution. Our method accommodates physical interactions with the environment without compromising safety. In a simulated wire harness assembly task, it achieved a 76\% success rate with no collisions or overextensions, outperforming prior methods that were either overly conservative or lacked DLO-specific safeguards. Future work will focus on closing the sim-to-real gap by validating the framework on physical hardware, improving contact modeling for more accurate tension prediction, and extending the approach to more complex tasks, such as multi-arm coordination or human-robot collaboration.

\bibliographystyle{IEEEtran}
\bibliography{references}



\end{document}